\documentclass{article}
\usepackage{spconf,amsmath,epsfig}

\usepackage{algorithmic}
\usepackage{algorithm}
\usepackage{amsmath}
\usepackage{amssymb}
\usepackage{graphicx}
\usepackage{multirow}
\usepackage{multicol}

\usepackage{graphicx}
\usepackage{xcolor}

\let\OLDthebibliography\thebibliography
\renewcommand\thebibliography[1]{
  \OLDthebibliography{#1}
  \setlength{\parskip}{0pt}
  \setlength{\itemsep}{0pt plus 0.3ex}
}

\begin{document}\sloppy

\def\x{{\mathbf x}}
\def\L{{\cal L}}

\title{FedNS: Improving Federated Learning for collaborative image classification on mobile clients}
%
\name{Yaoxin Zhuo, Baoxin Li}
\address{CIDSE, Arizona State University\\
\{yzhuo6, baoxin.li\}@asu.edu
}

\maketitle

\begin{abstract}
Federated Learning (FL) is a paradigm that aims to support loosely connected clients in learning a global model collaboratively with the help of a centralized server. 
The most popular FL algorithm is Federated Averaging (FedAvg), which is based on taking weighted average of the client models, with the weights determined largely based on dataset sizes at the clients. 
In this paper, we propose a new approach, termed Federated Node Selection (FedNS), for the server's global model aggregation in the FL setting. FedNS filters and re-weights the clients' models at the node/kernel level, hence leading to a potentially better global model by fusing the best components of the clients. Using collaborative image classification as an example, we show with experiments from multiple datasets and networks that FedNS can consistently achieve improved performance over FedAvg. 
\end{abstract}

\section{Introduction}

Mobile devices with cameras have become ubiquitous and they potentially produce the amount of data necessary for training deep learning models that have seen increasing adoption in applications such as computer vision and robotics. In most real applications, it is not realistic to expect that a centralized server may train a global model by collecting data from all involved mobile clients, due to bandwidth constraint (e.g., too much traffic between the server and millions of clients) and/or privacy concerns (e.g., users may not be willing to share images from their devices). The {\it Federated Learning} \cite{mcmahan2017communication} was proposed and advocated as a good strategy to address these issues, which allows loosely connected clients to contribute to the learning of a global model collaboratively.

In a typical FL setup, training data are distributed over clients, each of which learns to update a global model using its local data. A server maintains and updates the server's global model by aggregating copies of the local model from all the clients. A commonly used algorithm, named {\it Federated Averaging} (FedAvg)  \cite{mcmahan2017communication}, updates the global model by a weighted average of the client models, where the weights are proportional to the amount of data used by each involved client (in real implementations, not all clients are involved in every update). 

While FedAvg delivered state-of-the-art results in some reports, it still has some limitations. First, its assumption about the data distributions across clients is often violated in real-world scenarios. Second, simple weighted average of client models may not be optimal, as analyzed in some recent work   \cite{xiao2020averaging}. There are also other variants of FL, such as CMFL  \cite{luping2019cmfl} and FedMA \cite{Wang2020Federated}, where typically additional constraints or processing steps are introduced (e.g., more hyper-parameters and more computations at the clients), making the method less generally applicable compared with the simpler FedAvg.

In this paper,  we propose a new approach, termed Federated Node Selection (FedNS), for the server's global model aggregation in the FL setting. The approach filters and re-weights the clients' models at the node/kernel level on the server, hence leading to a potentially better global model by fusing the best components of the clients. We focus on a better way of module aggregation on the server while keeping the simplicity of the FL paradigm. Using collaborative image classification as a case study, we demonstrate the benefits of the proposed method from the following perspectives. First, the learning paradigm follows the FL framework without extra computation requirement for the clients or extra communication burden between the clients and the server. Second, our method can handle iid or non-iid  distributions without pre-defined hyper-parameters. Third, our method allows the clients to have varying amount of data, making it closer to real-world scenarios (where mobile clients could see and process dynamically changing data everyday). Experiments with three different datasets and networks are used to assess the performance of the proposed method. 

The remainder of this paper is organized as follows. In Section \ref{section related work}, we review related work and discuss the disadvantages wherever applicable. In Section \ref{section method}, we first introduce common baselines and then present our method FedNS. We design and perform experiments in Section \ref{section exp design} for evaluation. 
And finally, we conclude with Section \ref{section conclusion}.


\section{Related work}
\label{section related work}

Since the initial paper by McMahan {\it et al.} \cite{mcmahan2017communication}, FL has received considerable attention and there have been numerous research papers on this regard, of which we can only briefly review some that are most relevant to our work. A couple of recent surveys \cite{kairouz2019advances, li2020survey, li2020federated} may provide an ample picture for an interested reader. 
%
One line of research has been on saving the communication between the clients and the server. In \cite{konevcny2016federated}, a compressed structure was designed to reduce the communication cost. Horvath {\it et al.} \cite{horvath2019natural} introduced another compression technique for lowering the training time. Alistarh {\it et al.}  \cite{alistarh2017qsgd} proposed a family of compression schemes (QSGD), allowing a user to trade off between bandwidth and communication rounds (and hence modifying the basic FL framework). Federated Dropout \cite{bouacida2020adaptive} was proposed to allow the clients to train on only a subset of the global model, reducing parameters that need to be communicated to the server. In general, as the focus is on saving communication, such methods typically do not improve the learned global model, if they do not cause degradation.

Another direction is on improving the effectiveness of FL. Wang {\it et al.} \cite{luping2019cmfl} proposed CMFL that picks a fraction of all communicated clients to update the global model. The main limitation is the need for searching optimal hyper-parameters (thresholds). Nishio {\it et al.}  \cite{nishio2019client} tried to do client-level selective aggregation, based on the client's computing capability. Zhao {\it te al.} \cite{zhao2018federated} designed a strategy for improving accuracy when every client has only a single class of data. In \cite{tuorovercoming}, a method was proposed to filter the local data at client. But the filtering is based on pre-trained models. Yao {\it et al.}  \cite{yao2019towards} proposed FedFusion, an approach that could improve the performance by fusing the two model features, although extra computation is needed for choosing the best fusion approach. FedMA \cite{Wang2020Federated} constructs the global model in a layer-wise manner, which appears to cause significant burden on the clients.

There are also some efforts on improving the FL for specific tasks/applications. For example, Huang {\it et al.}  \cite{huang2019patient} designed a community-based FL algorithm that clusters distributed patient data into meaningful communities. 
Jennny {\it et al.} \cite{hamer2020fedboost} designed an approach for reducing the cost of communication, focusing on enabling a large model to be learned on relatively small memory of the clients.

While some of the aforementioned methods improve upon the more foundational FedAvg approach under peculiar settings, they often do not readily apply to more general FL settings, owing to the assumptions they introduced to constrain the problem. Our objective of this research is to design a method that focus on better model aggregation with minimal modification to the basic FL framework. For example, we relax the assumption about the knowledge of the data distribution. Hence our method has a better potential for wider applicability, especially for real-world applications.


\section{Proposed Method}
\label{section method}
In this section, we begin by introducing the baseline method of FL: Federated Averaging (FedAvg) in Section  \ref{subsection FedAvg}. Then we present a straightforward extension in Section \ref{subsection FedAvg+lastFC}, termed FedAvg+lastFC, which attempts to do a better weighted averaging for the last FC layer under the non-iid condition. We propose our FedNS method in Section \ref{subsection FedNS}, which is an improvement over FedAvg+lastFC in that better weighted averaging is done for all layers.

\subsection{\textbf{The FedAvg Baseline}}
\label{subsection FedAvg}
FedAvg was first proposed in \cite{mcmahan2017communication}, where the problem was formulated as one of optimizing the following non-convex neural network objective function: 
\begin{eqnarray}
    f(w)&=\frac{1}{K}\sum_{k=1}^K \ell(x_k,y_k;w_k)
\end{eqnarray}
where $k$ is the client index ranging from $1$ to $K$. 

The loss function $\ell(x_k,y_k;w_k)$ is the  $k$-th client's loss, i.e., the loss of model $w_k$ under data $x_k$ with corresponding label $y_k$. In the first step of FedAvg, the server initializes a global model $w_0$ and sends it to a fraction of communicated clients. Each communicated client $k$ will train the model with its local data, i.e., calculating the local gradient $g_k^i$ and updating the local model after training certain $E$ epochs (assuming with the learning rate $\eta$):
\begin{eqnarray}
w_{k,t}^{i+1} = w_{k,t}^i - \eta g_k^i
\end{eqnarray}
where $t$ indexes the communication rounds, ranging from $1$ to $N$, and $i$ is the client's local epoch range from $1$ to $E$. 

The initial model of the client is the weights received from the server. The clients will upload their updated local models back to the server. The server will then aggregate models from all clients using a  weighted averaging scheme:
\begin{eqnarray}
\label{ave equation}
    w_{s, t+1} = \sum_{k=1}^K \frac{n_k}{n} w_{k, t}
\end{eqnarray}
where $w_s$ is the weights of the global model on the server, $n_k$ is the total number of training samples on client $k$ and $n$ is the sum of all samples from all communicated clients. It is worth noting that Eqn. \ref{ave equation} will become simple averaging when all clients have the same amount of data. In the next communication round, the server will start from sending aggregated server model $w_{s,t}$ to all the clients and repeat the previous steps until reach the target communication rounds. The complete pseudo-code for this baseline is given in Algorithm\ref{alg:FedAvg}.

\begin{algorithm}[t]
    \caption{Federated Averaging (FedAvg)}
    \label{alg:FedAvg}
    \begin{algorithmic}
    \STATE{\textbf{Server executes:}}
        \STATE initialize $w_0$ as $w_{s,1}$
            \FOR{each communication round $t$  from $1$ to $N$}
                \STATE{$S_t \leftarrow$ (the fraction of communicated clients)}
                \FOR{each client $k \in S_t$ \textbf{in parallel}}
                    \STATE{$w_{k, t} \leftarrow $ ClientUpdate($k, w_{s,t}$) }
                \ENDFOR
                \STATE{\textbf{Model Aggregation:}}
                \STATE{$w_{s, t+1} \leftarrow \sum_{k=1}^K \frac{n_k}{n} w_{k,t}$}
            \ENDFOR
    \STATE
    \STATE{\textbf{ClientUpdate($k,w_{k,t}$):}}
        \STATE $\mathcal{B}\leftarrow$ (split local data into batches of local batch size $B$)
            \FOR{each local epoch $i$ from $1$ to $E$}
                \FOR{batch $b \in \mathcal{B} $}
                    \STATE{$g_k^i = \triangledown (w_{k,t}^i;b)$}
                    \STATE{$w_{k,t}^{i+1} \leftarrow w_{k,t}^i - \eta g_k^i $}
                \ENDFOR
            \ENDFOR
            \STATE{return $w_{k,t}$ to server}
    \end{algorithmic}
\end{algorithm}

\subsection{\textbf{FedAvg+lastFC: A Simple Extension to FedAvg}}
\label{subsection FedAvg+lastFC}
One may build a naive and straightforward extension to FedAvg for potential improvement, by using weights based on the number of samples for each class (instead of the number of samples) at each client when doing the weighted averaging for the last fully connected (FC) layer. That is, we define 
\begin{eqnarray}
   w_{s, t+1}^{l=L,c} = \sum_{k=1}^K \frac{n_k^c}{n^c} w_{k,t}^{l=L,c}
\end{eqnarray}
where $l$ is the index of layers in a model that has $L$ layers, 
$c$ is the index for the last FC layer's hidden nodes, ranging from $1$ to $C$. And $C$ is also equals to the total number of classes in image classification models. So $n_k^c$ is client $k$'s number of samples in the class that corresponding to node $c$ and $n^c$ is the total number of data in class $c$ within all communicated clients. This leads to Algorithm \ref{alg:FedAvg+lastFC}, where only the model aggregation pseudo-code is given to save space.

\begin{algorithm}[t]
    \caption{Federated Averaging with weighted averaging last FC layer (FedAvg+lastFC)}
    \label{alg:FedAvg+lastFC}
    \begin{algorithmic}
    \STATE{\textbf{Server executes:}}
        \STATE{......}
                \STATE{\textbf{Model Aggregation:}}
                \FOR{each layer $l$ from $1$ to $L$}
                    \IF{$l<N$}
                        \STATE{$w_{s, t+1} \leftarrow \sum_{k=1}^K \frac{n_k}{n} w_{k,t}$}
                    \ELSE
                        \FOR{each node $c$ in the layer $L$ from $1$ to $C$}
                            \STATE{$w_{s, t+1}^{l=L,c} = \sum_{k=1}^K \frac{n_k^c}{n^c} w_{k,t}^{l=L,c}$}
                        \ENDFOR
                    \ENDIF
                \ENDFOR
        \STATE{......}
    \end{algorithmic}
\end{algorithm}

\subsection{FedNS: the Proposed Method}
\label{subsection FedNS}
While FedAvg+lastFC employs an intuitive way of weighted averaging for the last FC layer, the method cannot be easily extended to previous layers since they lack explicit correspondence to the number of classes. In the meantime, using only the number of samples at each client for weighted averaging is not necessarily optimal. Hence we propose to use the variance of the weight update to evaluate the contribution of a node/kernel to the client's local model (for layers other than the last FC layer). During the aggregation stage, the server could easily calculate the differential weights for each node and calculate its variance (and derive from this weights for the clients) since it has the global model from the previous communication round. We term this approach Federated Node Selection (FedNS). 
It first calculates the following: 
\begin{eqnarray}
   v_{k,t+1}^{l\neq L,c} = \text{variance}(w_{k,t+1}^{l\neq L,c} - w_{k,t}^{l\neq L,c})
\end{eqnarray}
where $c$ is the node index ranges from $1$ to $C$. $C$ is the total node/kernel number for the layer $l$ in a model that has $L$ layers. We calculate their mean ($\mu_{t+1}^{l\neq L,c}$) and sigma ($\sigma_{t+1}^{l\neq L,c}$) after getting all the variances. If the node's variance is out of range $[\mu_{t+1}^{l\neq L,c}-2\sigma_{t+1}^{l\neq L,c}, \mu_{t+1}^{l\neq L,c}+2\sigma_{t+1}^{l\neq L,c}]$ ($95\%$ of confidence interval if the underlying distribution is normal), we will remove that node in the global model aggregation step. Then we re-normalize the remaining nodes' variances and get the new weights. The server do a new weighted averaging for all layers other than the last FC layer at node/kernel level by the new weights in the following equation: 
\begin{eqnarray}
   w_{s, t+1}^{l\neq L,c} = \sum_{k=1}^K \frac{ v_{k, t}^{l\neq L,c} }{ v_{t}^{l\neq L, c} } w_{k,t}^{l\neq L,c}
\end{eqnarray}
where $c$ is the remaining node's index, $v_{k, t}^{l\neq L,c}$ is node $c$'s variance of differential weights. And $v_{t}^{l\neq L, c}$ is the sum of variances from all remaining nodes. This new weighted averaging method leads to Algorithm \ref{alg:FedNS}, where only the model aggregation pseudo-code is given to save space.
\begin{algorithm}[t]
    \caption{Federated Node Selection (FedNS)}
    \label{alg:FedNS}
    \begin{algorithmic}
    \STATE{\textbf{Server executes:}}
        \STATE{......}
            \STATE{\textbf{Model Aggregation:}}
                \FOR{each layer $l$ from $1$ to $L$}
                    \IF{$l < L$}
                        \FOR{each node $c$ in the layer $l$ from $1$ to $C$}
                            \STATE{$v_{k,t}^{l\neq L,c} = \text{variance}(w_{k,t}^{l\neq L,c} - w_{k,t-1}^{l\neq L,c})$}
                            \STATE{\text{filter nodes and re-normalize remaining variances} }
                        \ENDFOR
                        \STATE{$w_{s,t+1}^{l\neq L,c} = \sum_{k=1}^K \frac{v_{k,t}^{l\neq L,c}}{v^c} w_{k,t}^{l\neq L,c}$}
                    \ELSE
                        \FOR{each node $c$ in the layer $L$ from $1$ to $C$}
                            \STATE{$w_{s, t+1}^{l=L,c} = \sum_{k=1}^K \frac{n_k^c}{n^c} w_{k,t}^{l=L,c}$}
                        \ENDFOR
                    \ENDIF
                \ENDFOR
        \STATE{......}
    \end{algorithmic}
\end{algorithm}


\section{Experiments}
\label{section exp design}
In this section, we present the empirical study of FedNS and comprehensively evaluate it. We compare FedNS with FedAvg and FedAvg+lastFC in two aspects: one is the final global model's performance and the other is every communication round's performance. The first one shows the model's performance after FL reaches the target communication rounds. And the second one shows which method is in a leading position during the FL process, which is also an important factor for real applications (e.g., whether a mobile client can still have good performance in a long process of incremental learning).

\subsection{Datasets}
To do the case study of collaborative image classification on mobile clients, we select three image classification datasets. Different from aforementioned methods that used simple MNIST \cite{lecun1998gradient} dataset, we replace it with a more challenging dataset: FashionMNIST \cite{xiao2017fashion}. The other two datasets we select are CIFAR10 \cite{krizhevsky2009learning} and tinyImageNet \cite{tinyImageNet}. Both FashionMNIST and CIFAR10 have 10 classes. FashionMNIST has $6,000$ training and $1,000$ testing images for each class. CIFAR10 has $5,000$ training and $1,000$ testing images for each class. TinyImageNet is a selective subset of ImageNet \cite{deng2009imagenet} and consist of 200 classes' images. It has $500$ training and $50$ testing images for each class.

\subsection{Networks}
We use CNN models of different complex levels for the datasets. For the FashionMNIST dataset, we use a simple CNN model that has 2 convolutional layers and 3 fully connected layers. The first $5\times5$ convolutional layer has 32 channels. The second $5 \times 5$ convolutional layer has 64 channels. Both of them are followed with $2\times 2$ ReLU activation and $2 \times 2$ max pooling. The 3 FC layers have 1024, 256 and 10 nodes respectively. All of them are followed by ReLU activation. For the CIFAR10 dataset, we use the AlexNet \cite{krizhevsky2012imagenet}. For the tinyImageNet dataset, we use the ResNet18 \cite{he2016deep}.

\subsection{Simulation Settings} 
Unlike some existing works that employed a simplistic settings of each client having  a fixed set of images throughout the entire training process, we allow each client to randomly choose a subset of images as local data in every communication round according to the iid or non-iid conditions. So the experiment setting is closer to real-world scenarios where mobile clients receive dynamically changing data. Specifically, for FashionMNIST, each client randomly chooses 5 images per class in the iid condition and 1-10 images per class in the non-iid condition; for CIFAR10, each client randomly chooses 50 images per class in the iid condition and 1-100 images per class in the non-iid condition; for tinyImageNet, each client randomly chooses 5 images per class in the iid condition and 1-10 images per class in the non-iid condition.

For other algorithmic parameters, we follow the same settings in \cite{mcmahan2017communication}. We fix the learning rate $\eta$ as $1e-2$. And we set communication round (day) $N=50$, local epoch $E=5$, local batch size $B=10$ and the communicated client number $K=10$, which means server could randomly communicate with $10$ clients in every communication round.

\subsection{Evaluation}
\label{section exp results}
For FashionMNIST and CIFAR10, we use accuracy, macro precision, macro recall and macro F-Score to evaluate the models. For tinyImageNet, we use top-1 accuracy and top-5 accuracy to evaluate the models. To visualize the performance difference in every communication round, we plot the differential performance numbers (against the FedAvg baseline). Because of the similar trend of accuracy and macro recall, we omit the macro recall numbers due to space limitation.

\subsection{Results}
\begin{figure}[t]
    \centering
    
    \begin{minipage}[b]{0.48\linewidth}
      \centering
      \centerline{\epsfig{figure=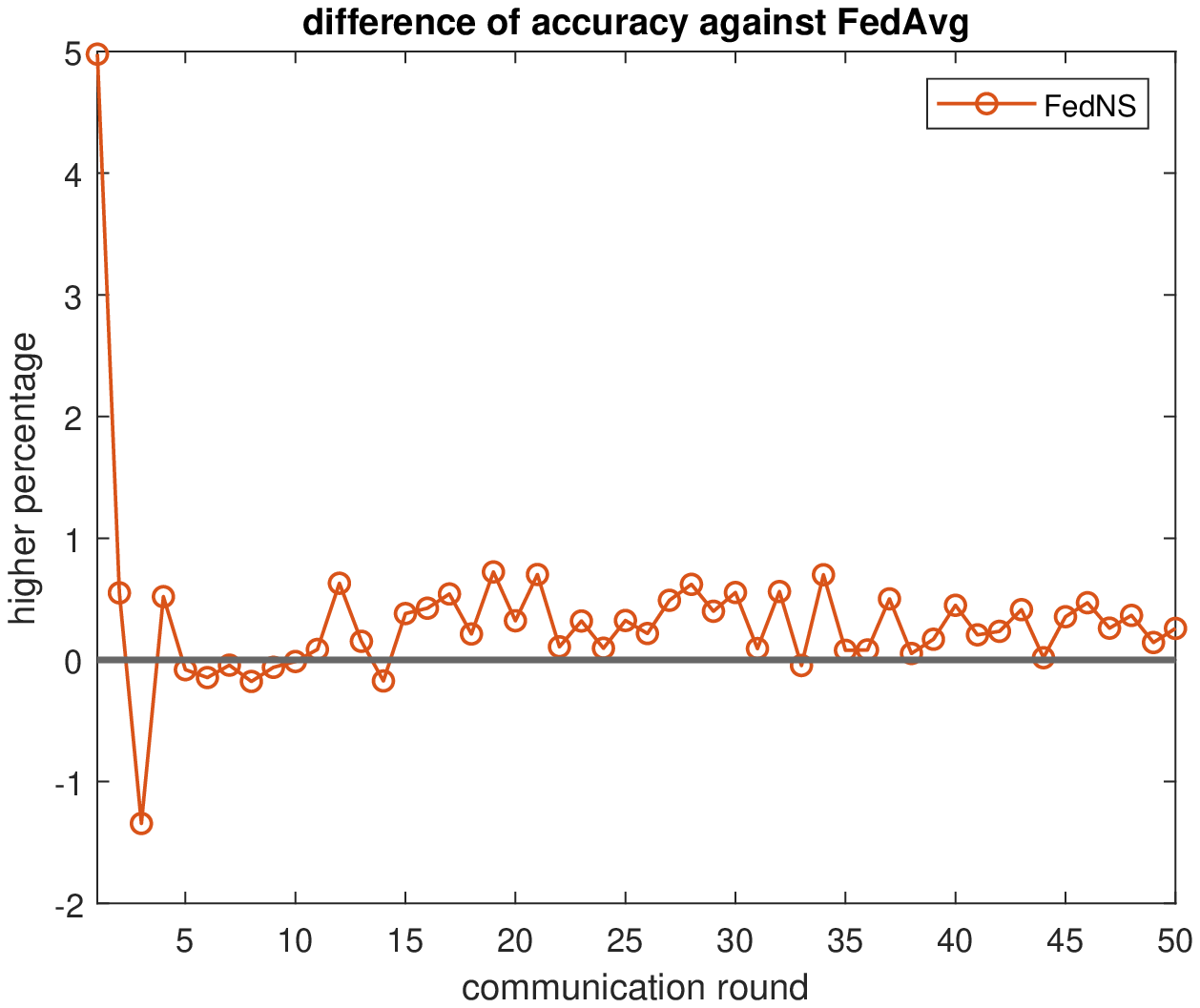,width=4cm}}
      \centerline{iid accuracy}\medskip
    \end{minipage}
    \begin{minipage}[b]{0.48\linewidth}
      \centering
      \centerline{\epsfig{figure=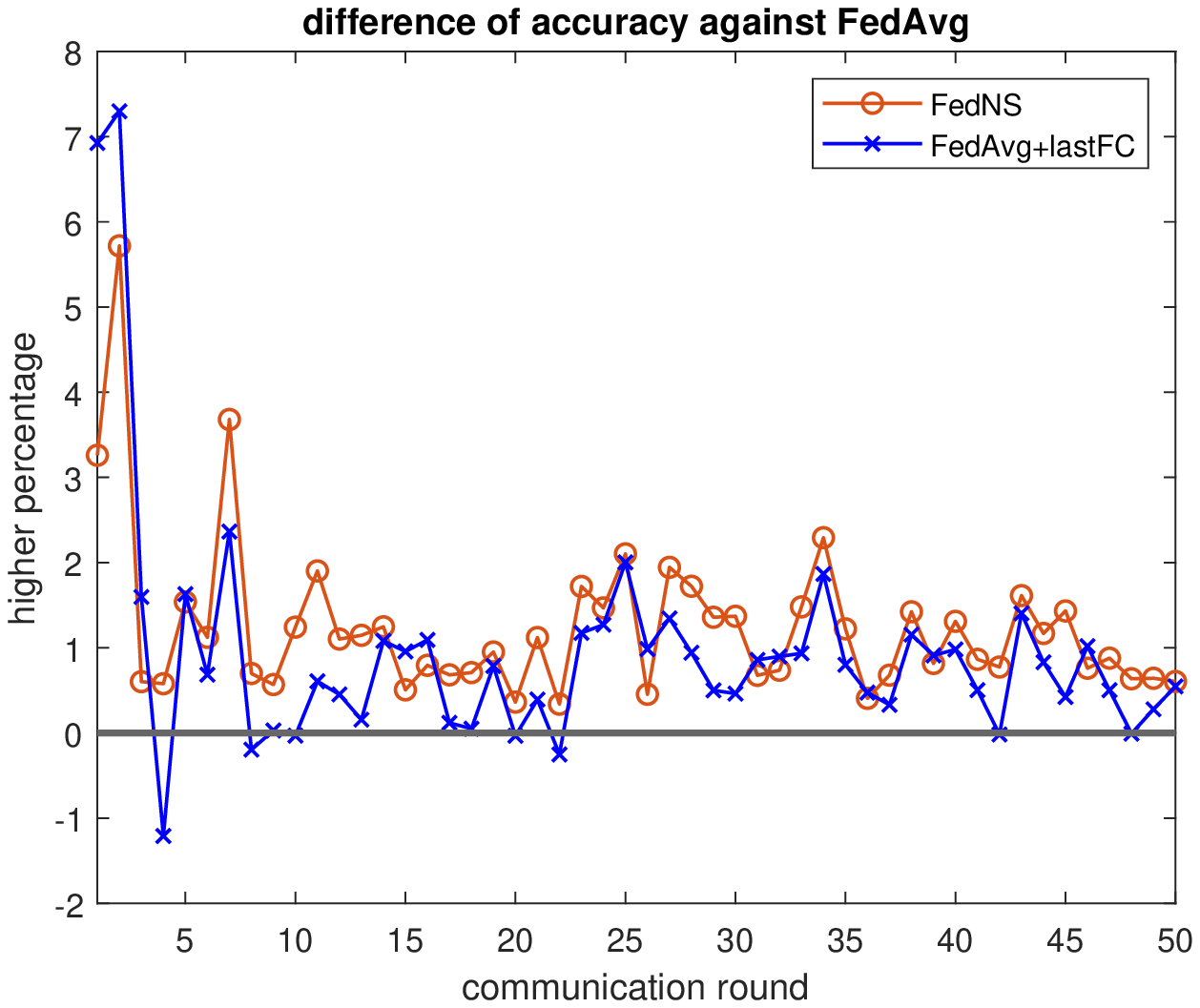,width=4cm}}
      \centerline{non-iid accuracy}\medskip
    \end{minipage}
    \begin{minipage}[b]{0.48\linewidth}
      \centering
      \centerline{\epsfig{figure=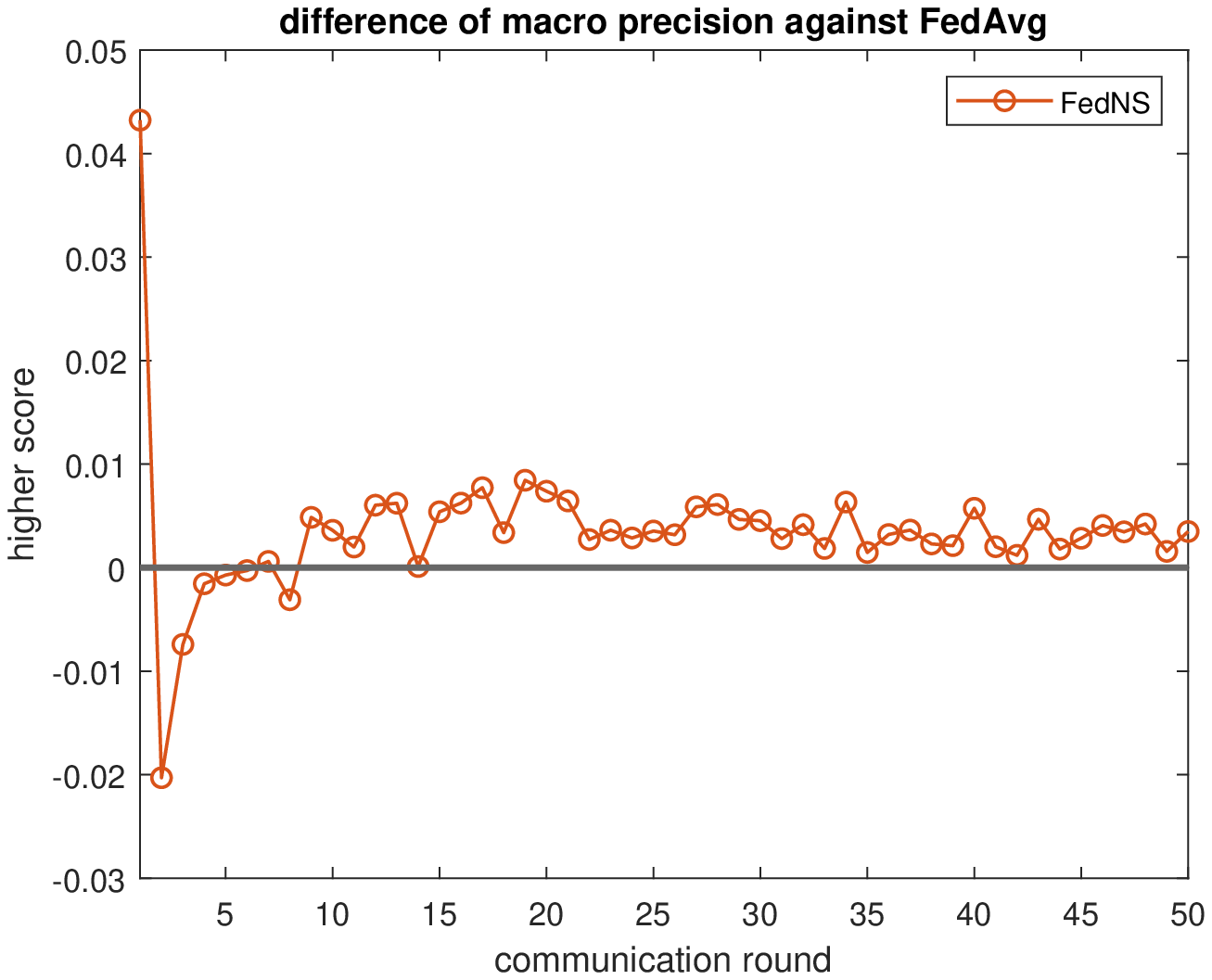,width=4cm}}
      \centerline{iid macro precision}\medskip
    \end{minipage}
    \begin{minipage}[b]{0.48\linewidth}
      \centering
      \centerline{\epsfig{figure=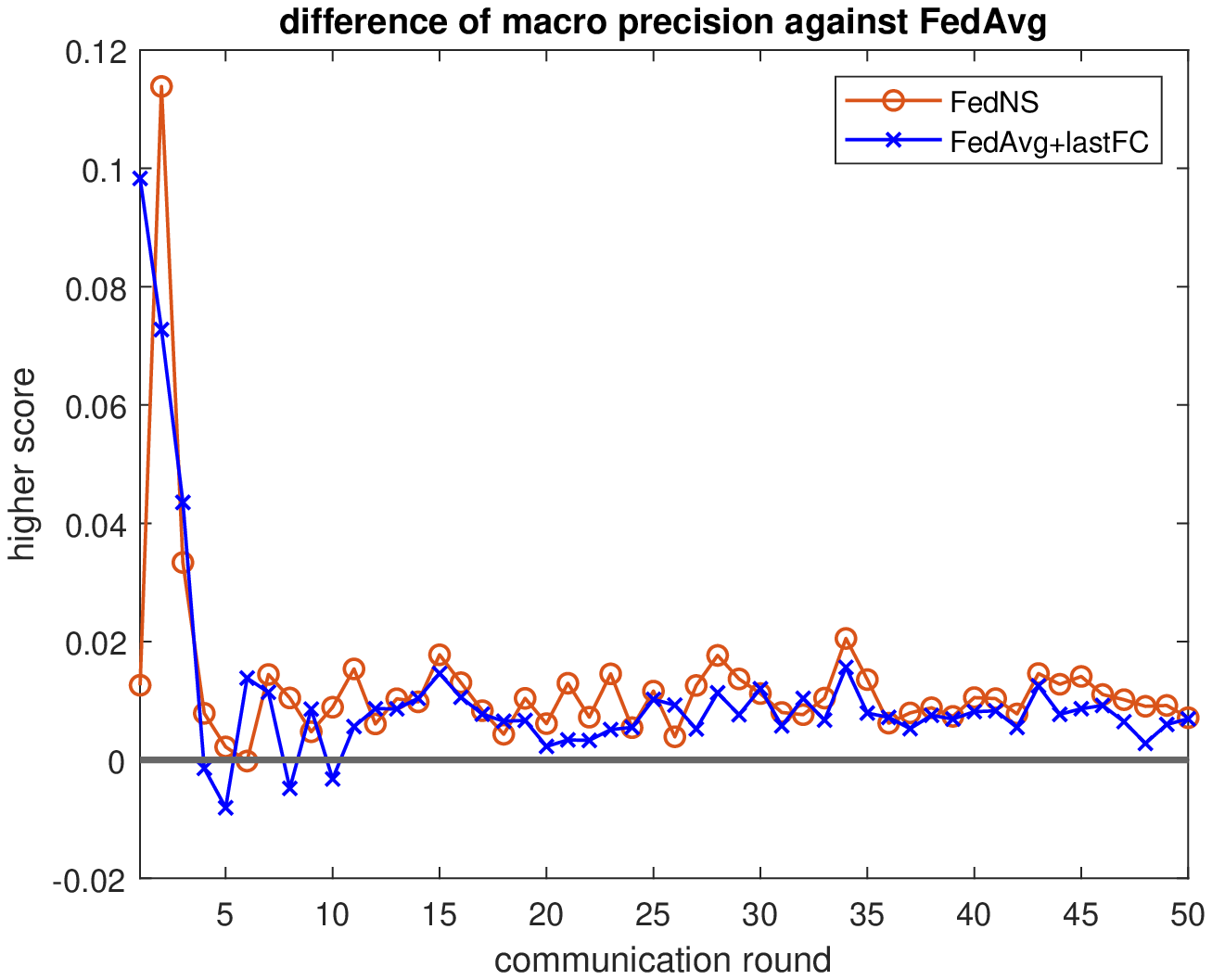,width=4cm}}
      \centerline{non-iid macro precision}\medskip
    \end{minipage}
    \begin{minipage}[b]{0.48\linewidth}
      \centering
      \centerline{\epsfig{figure=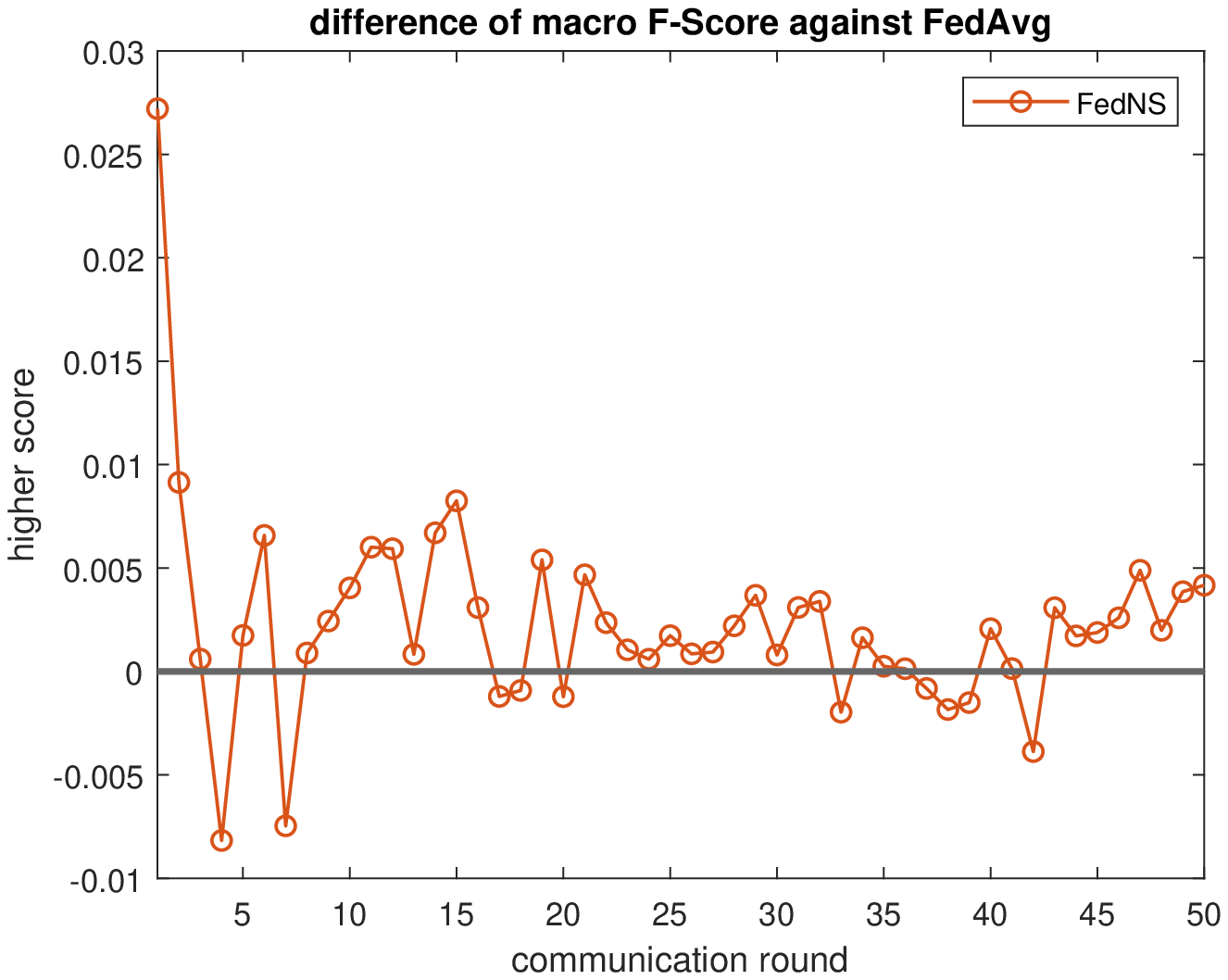,width=4cm}}
      \centerline{iid macro F-Score}\medskip
    \end{minipage}
    \begin{minipage}[b]{0.48\linewidth}
      \centering
      \centerline{\epsfig{figure=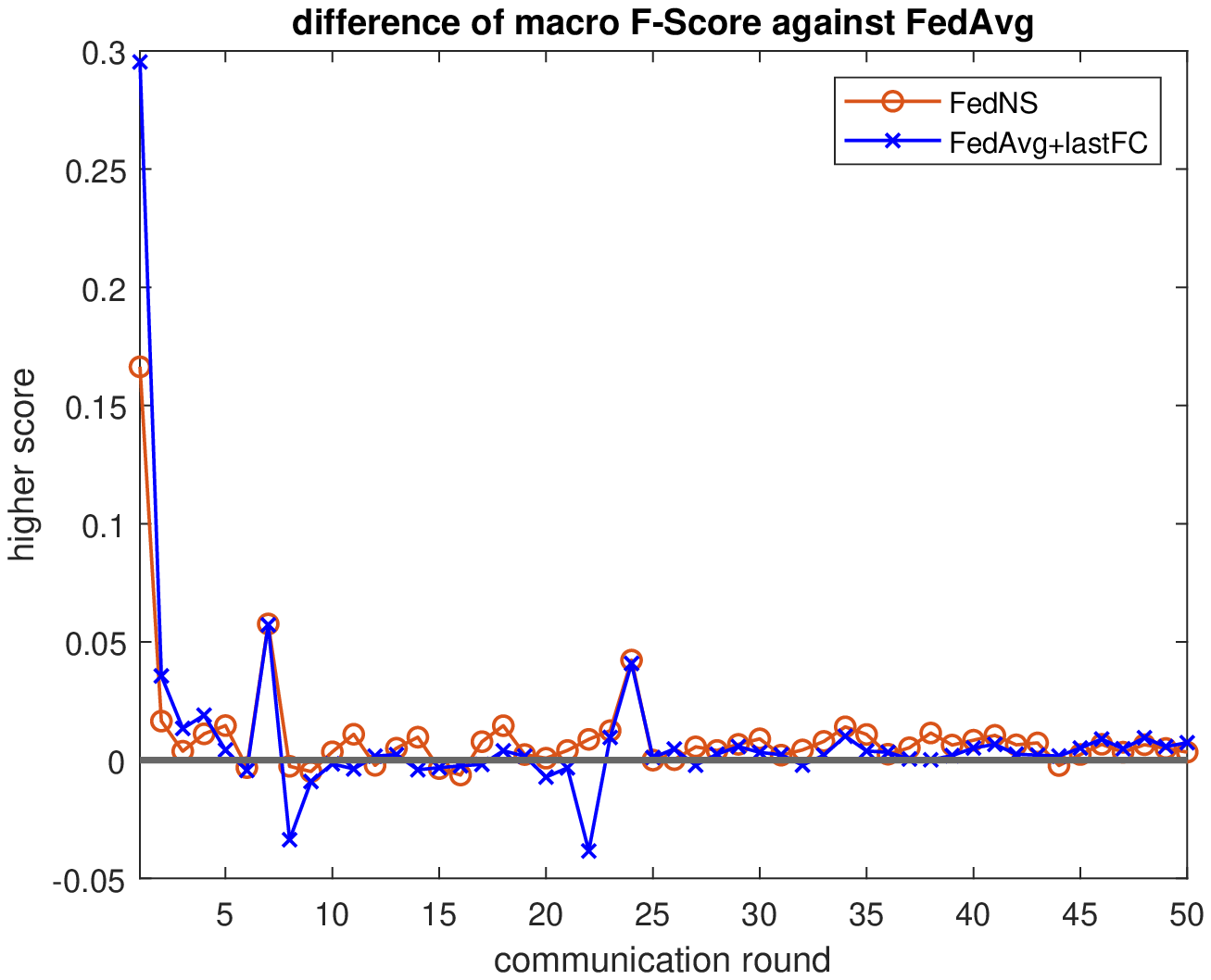,width=4cm}}
      \centerline{non-iid macro F-Score}\medskip
    \end{minipage}   

    \caption{FashionMNIST iid and non-iid daily comparison}
    \label{fig:FashionMNIST daily}
\end{figure}

\begin{figure}[t]
    \centering
    \begin{minipage}[b]{0.48\linewidth}
      \centering
      \centerline{\epsfig{figure=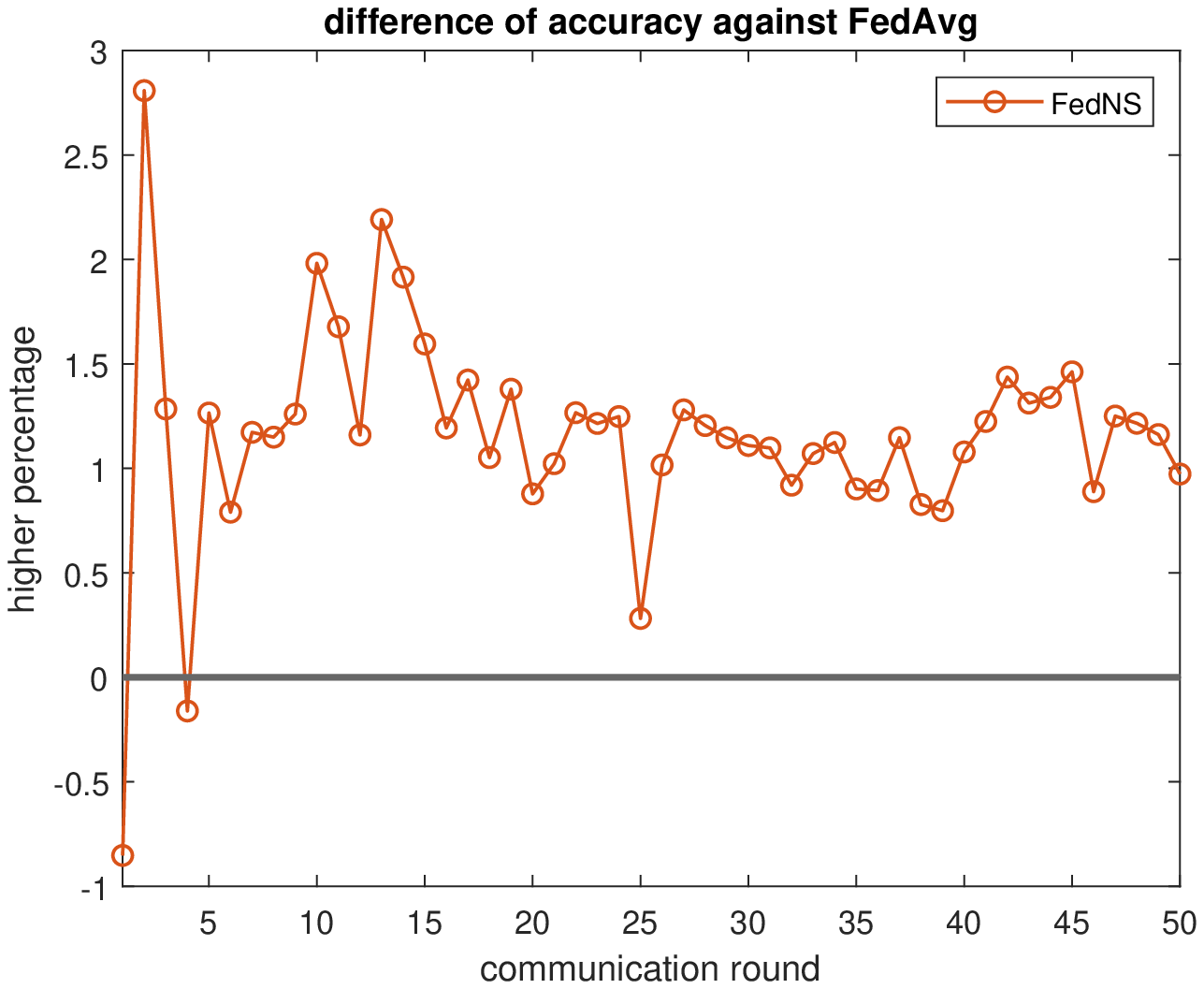,width=4cm}}
      \centerline{iid accuracy }\medskip
    \end{minipage}
    \begin{minipage}[b]{0.48\linewidth}
      \centering
      \centerline{\epsfig{figure=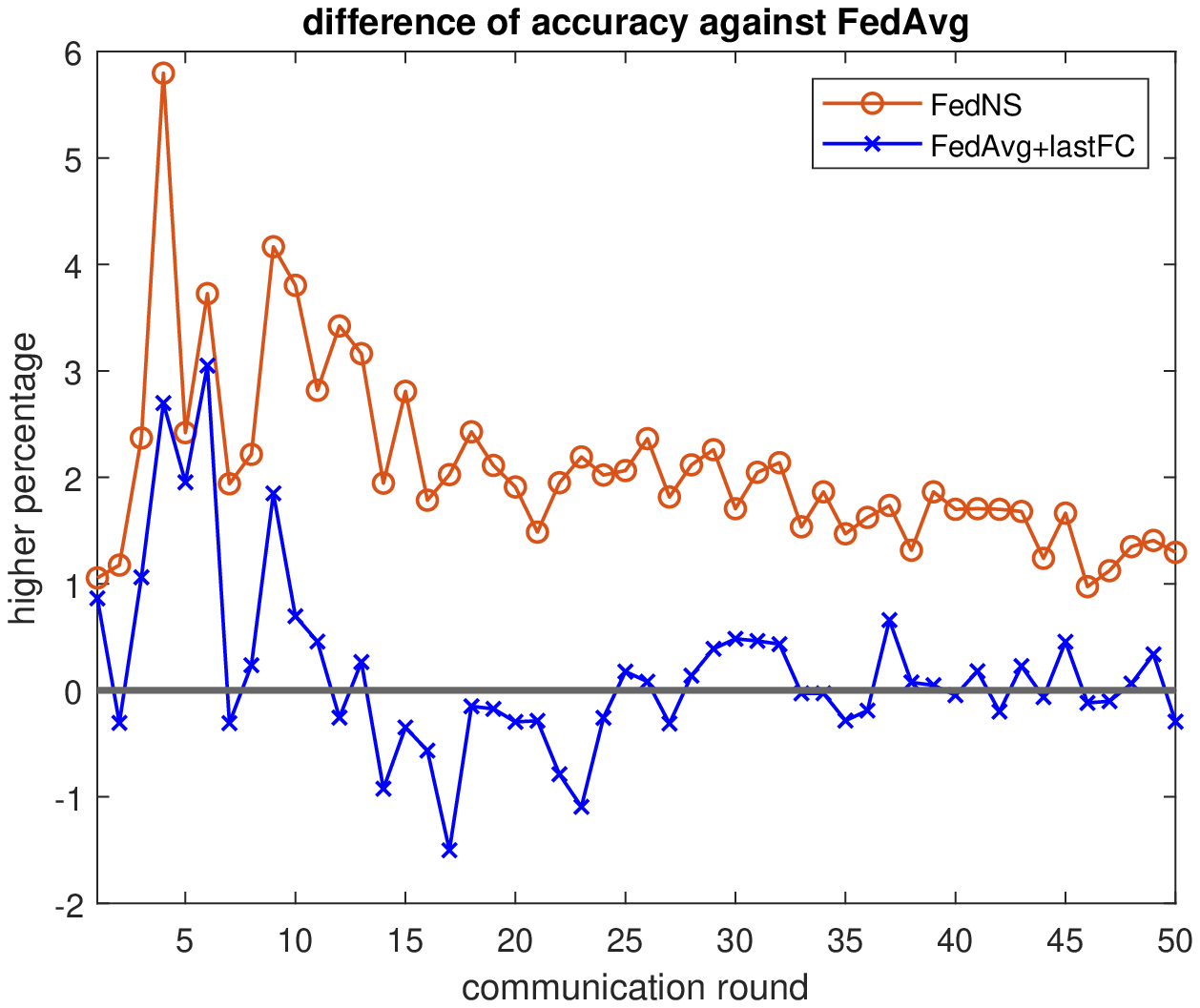,width=4cm}}
      \centerline{non-iid accuracy }\medskip
    \end{minipage}
    \begin{minipage}[b]{0.48\linewidth}
      \centering
      \centerline{\epsfig{figure=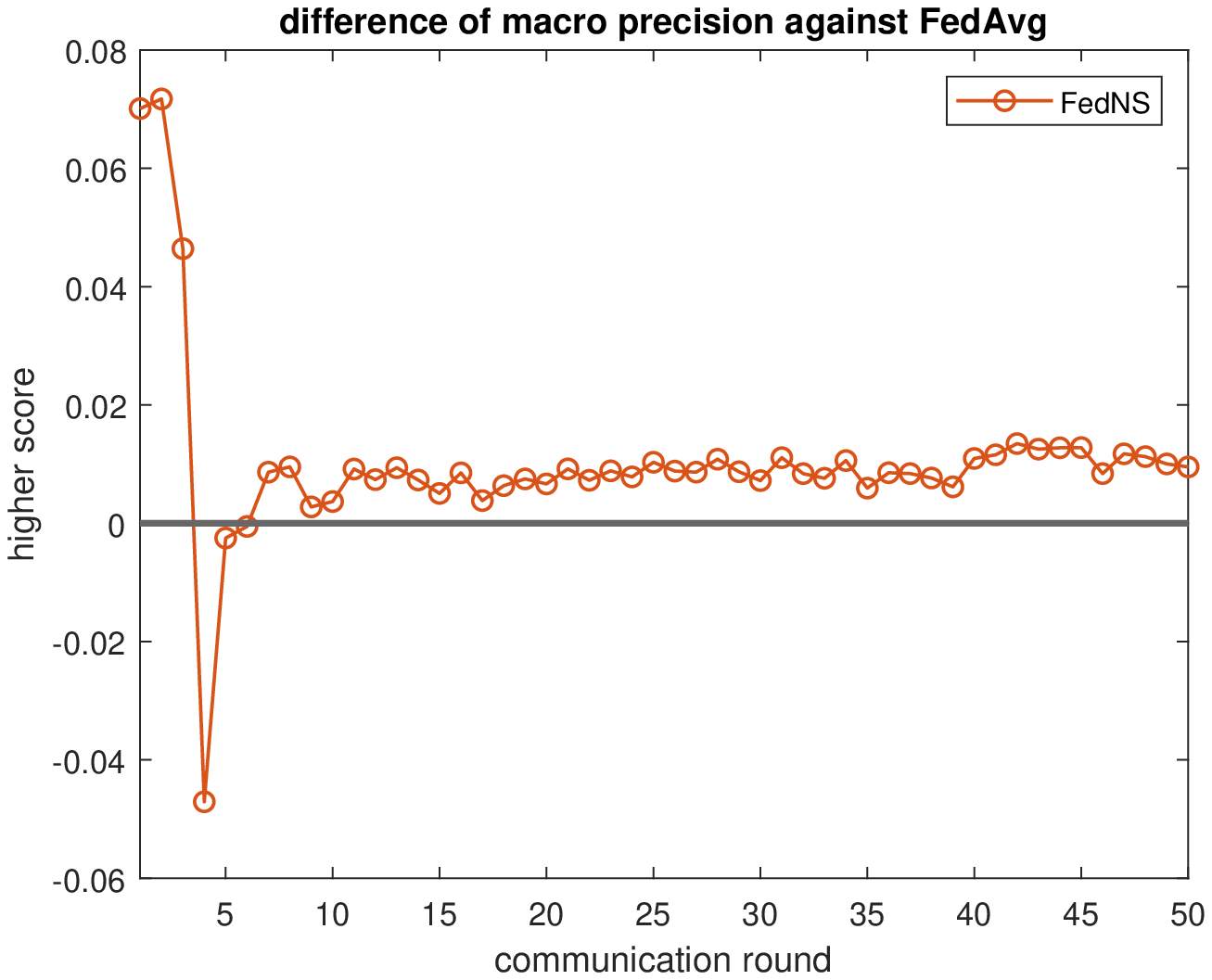,width=4cm}}
      \centerline{iid macro precision }\medskip
    \end{minipage}
    \begin{minipage}[b]{0.48\linewidth}
      \centering
      \centerline{\epsfig{figure=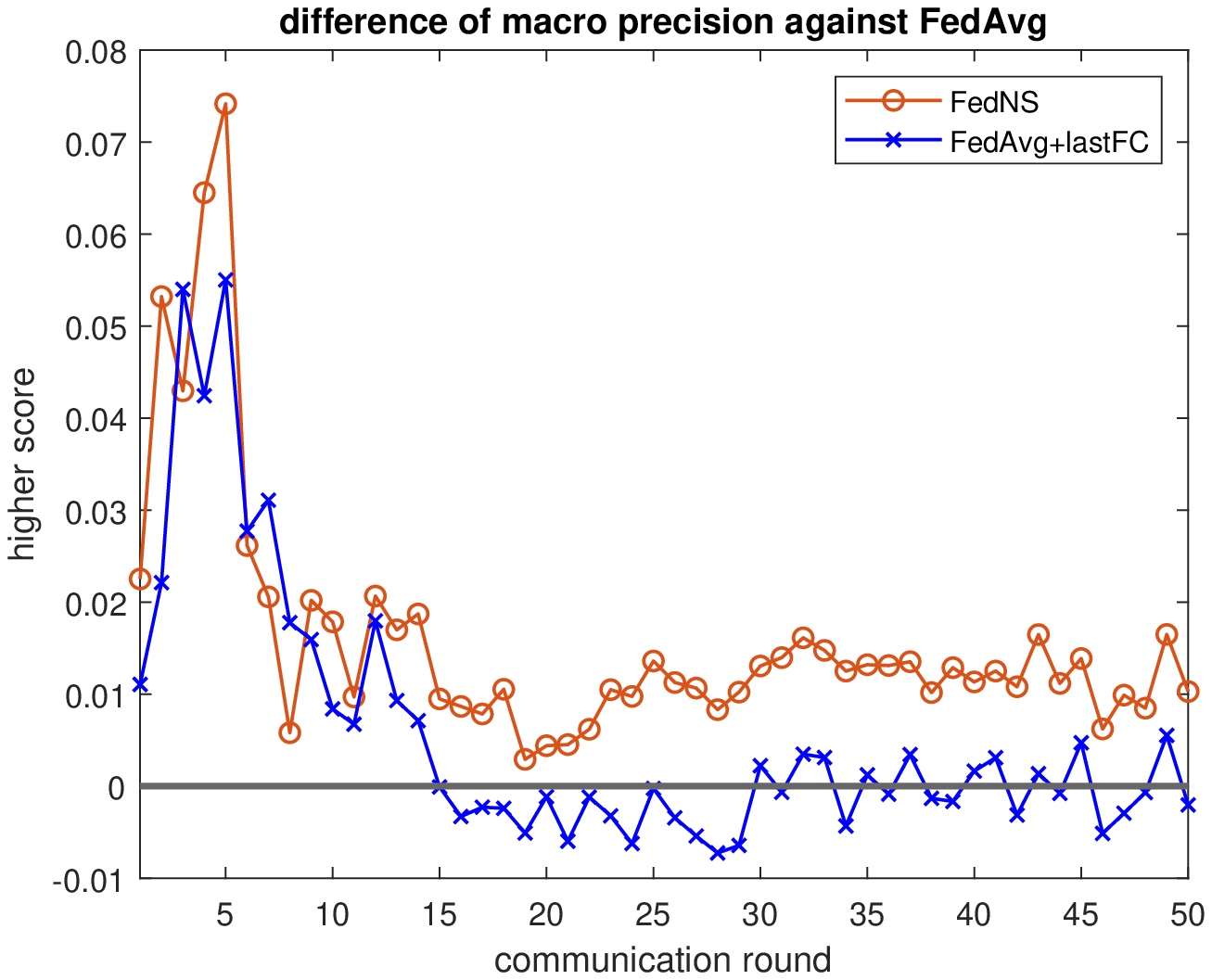,width=4cm}}
      \centerline{non-iid macro precision }\medskip
    \end{minipage}
    \begin{minipage}[b]{0.48\linewidth}
      \centering
      \centerline{\epsfig{figure=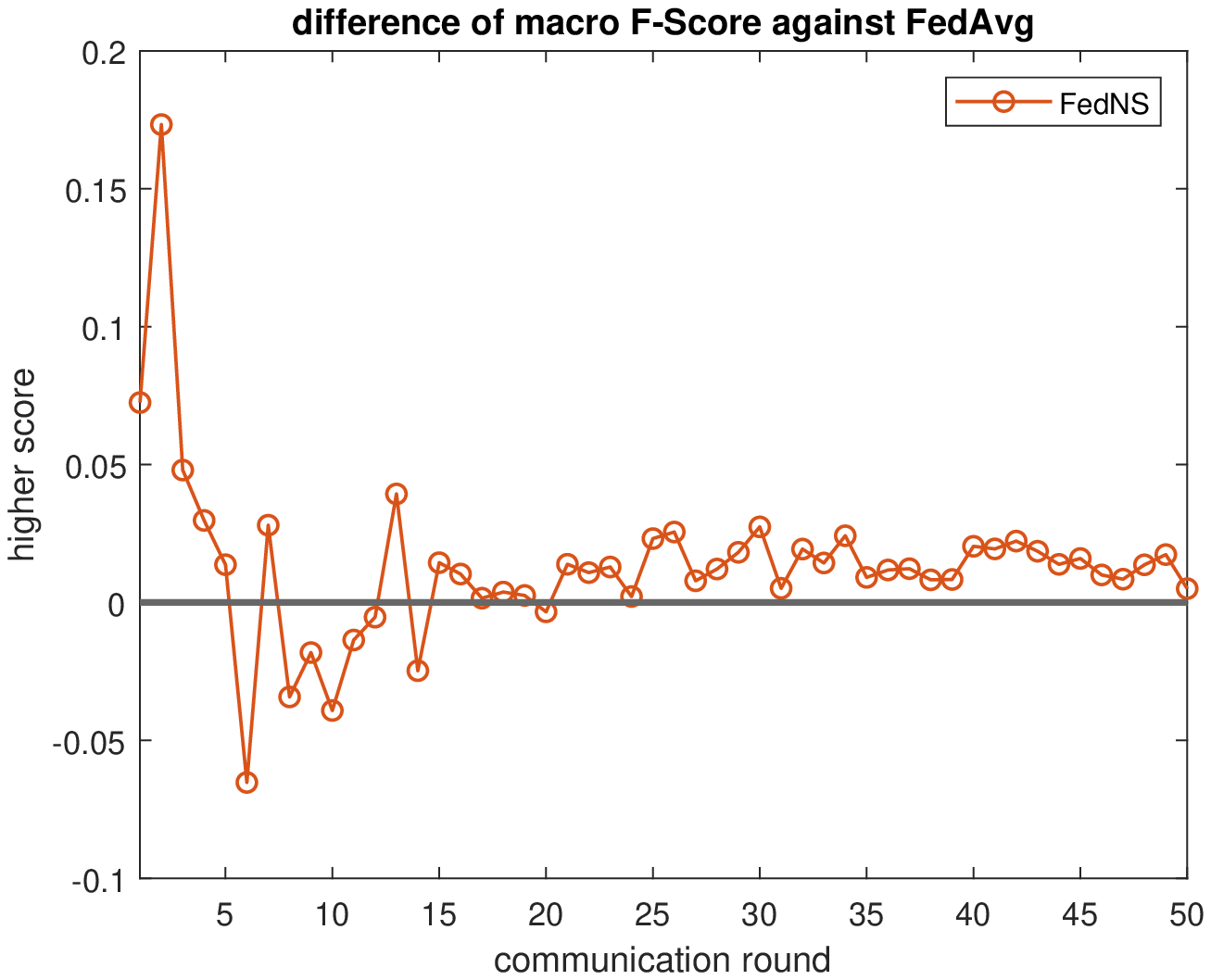,width=4cm}}
      \centerline{iid macro F-Score }\medskip
    \end{minipage}
    \begin{minipage}[b]{0.48\linewidth}
      \centering
      \centerline{\epsfig{figure=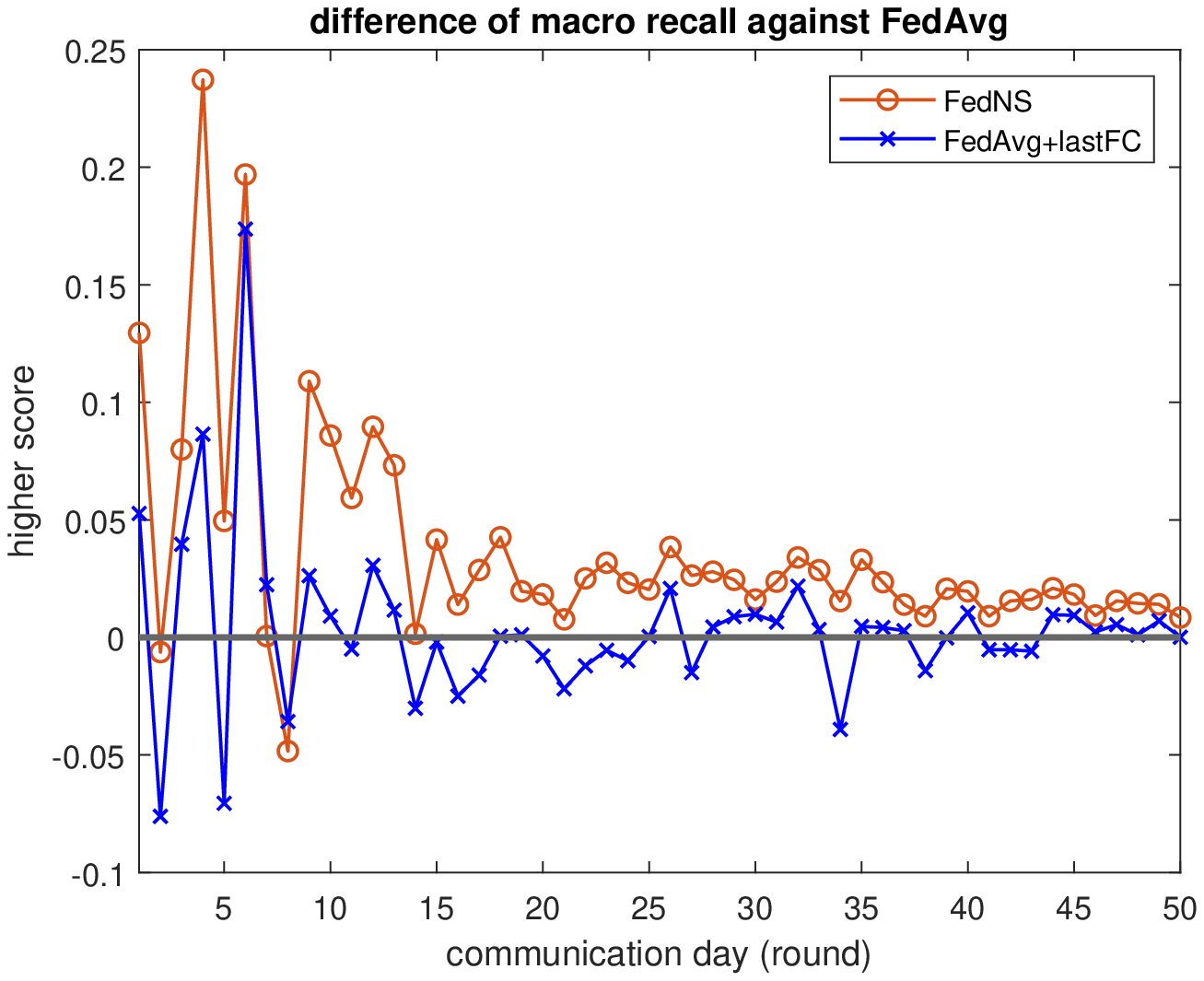,width=4cm}}
      \centerline{non-iid macro F-Score }\medskip
    \end{minipage}   
    
    \caption{CIFAR10 iid and non-iid daily comparison}
    \label{fig:CIFAR10 daily}
\end{figure}

\begin{figure}[t]
    \centering
    
    \begin{minipage}[b]{0.48\linewidth}
      \centering
      \centerline{\epsfig{figure=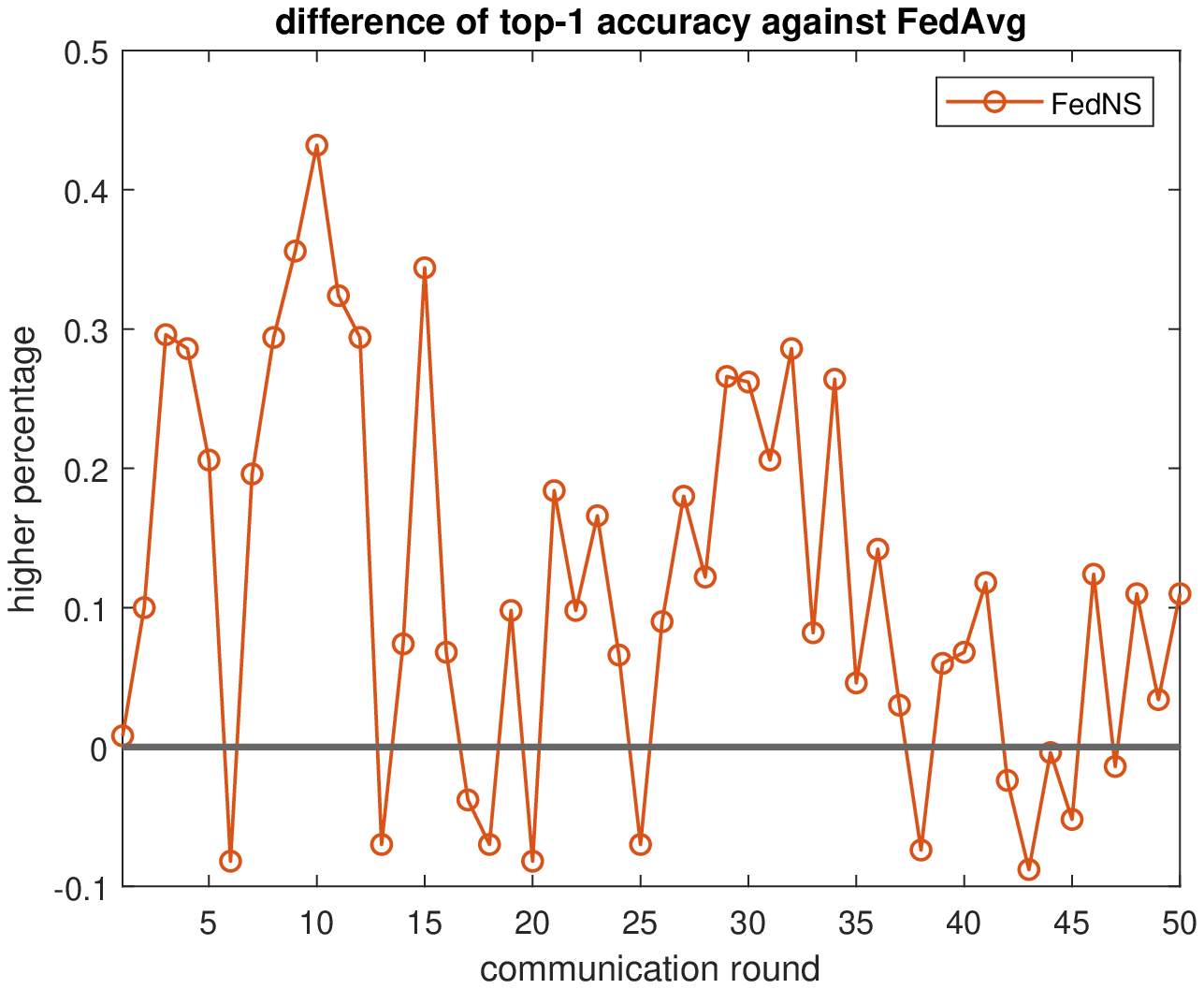,width=4cm}}
      \centerline{iid top-1 accuracy}\medskip
    \end{minipage}
    \begin{minipage}[b]{0.48\linewidth}
      \centering
      \centerline{\epsfig{figure=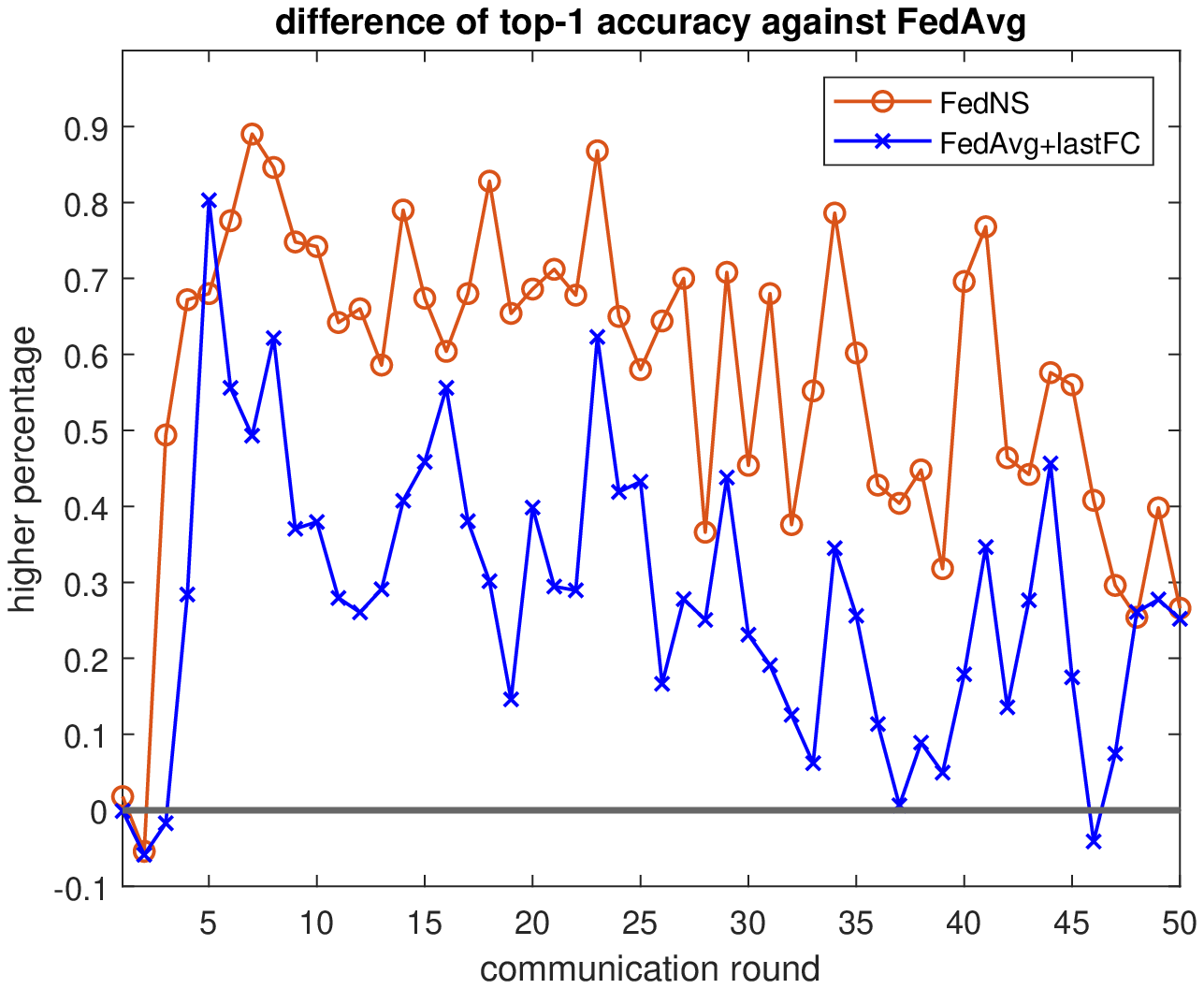,width=4cm}}
      \centerline{non-iid top-1 accuracy}\medskip
    \end{minipage}
    \begin{minipage}[b]{0.48\linewidth}
      \centering
      \centerline{\epsfig{figure=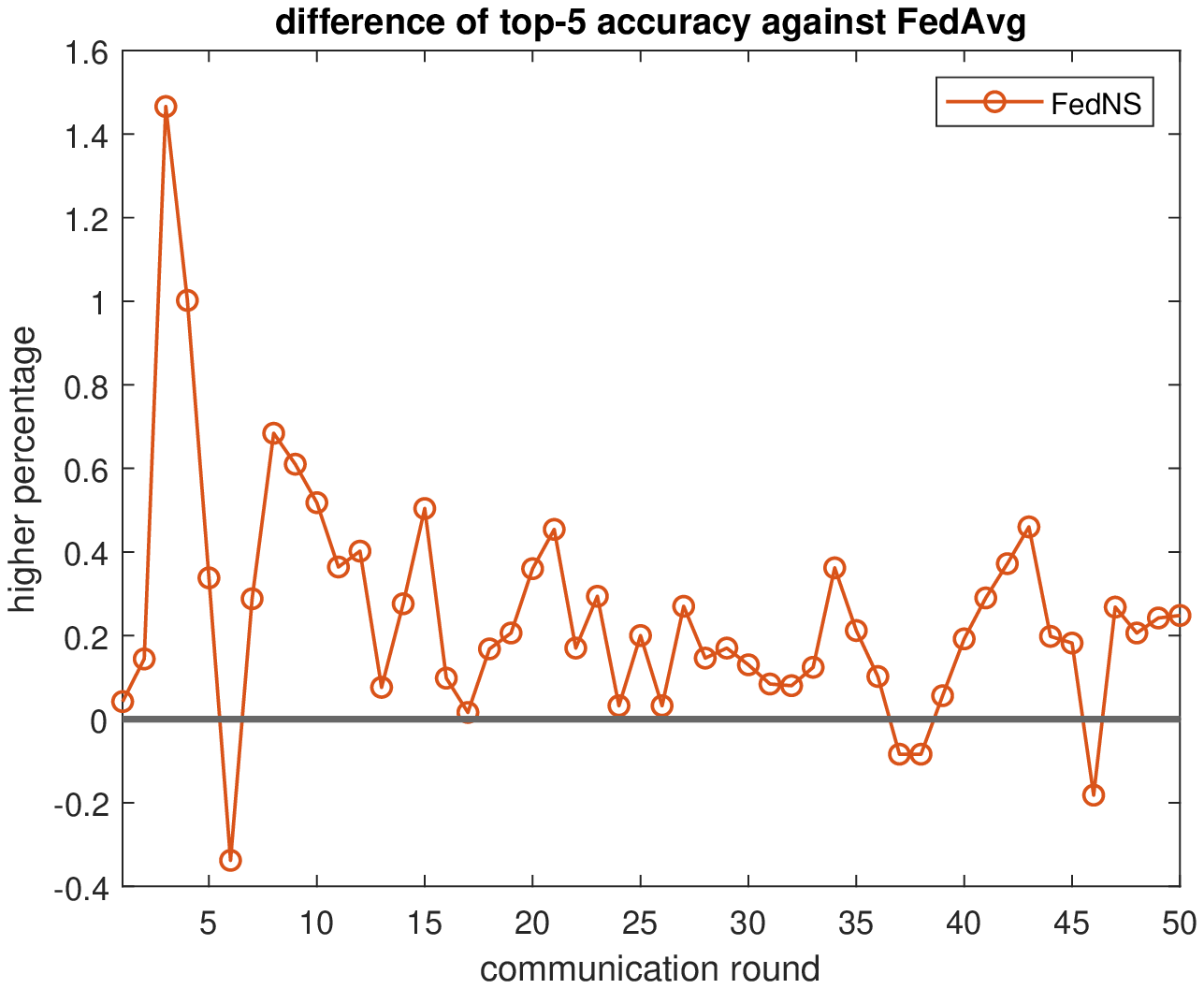,width=4cm}}
      \centerline{iid top-5 accuracy}\medskip
    \end{minipage}
    \begin{minipage}[b]{0.48\linewidth}
      \centering
      \centerline{\epsfig{figure=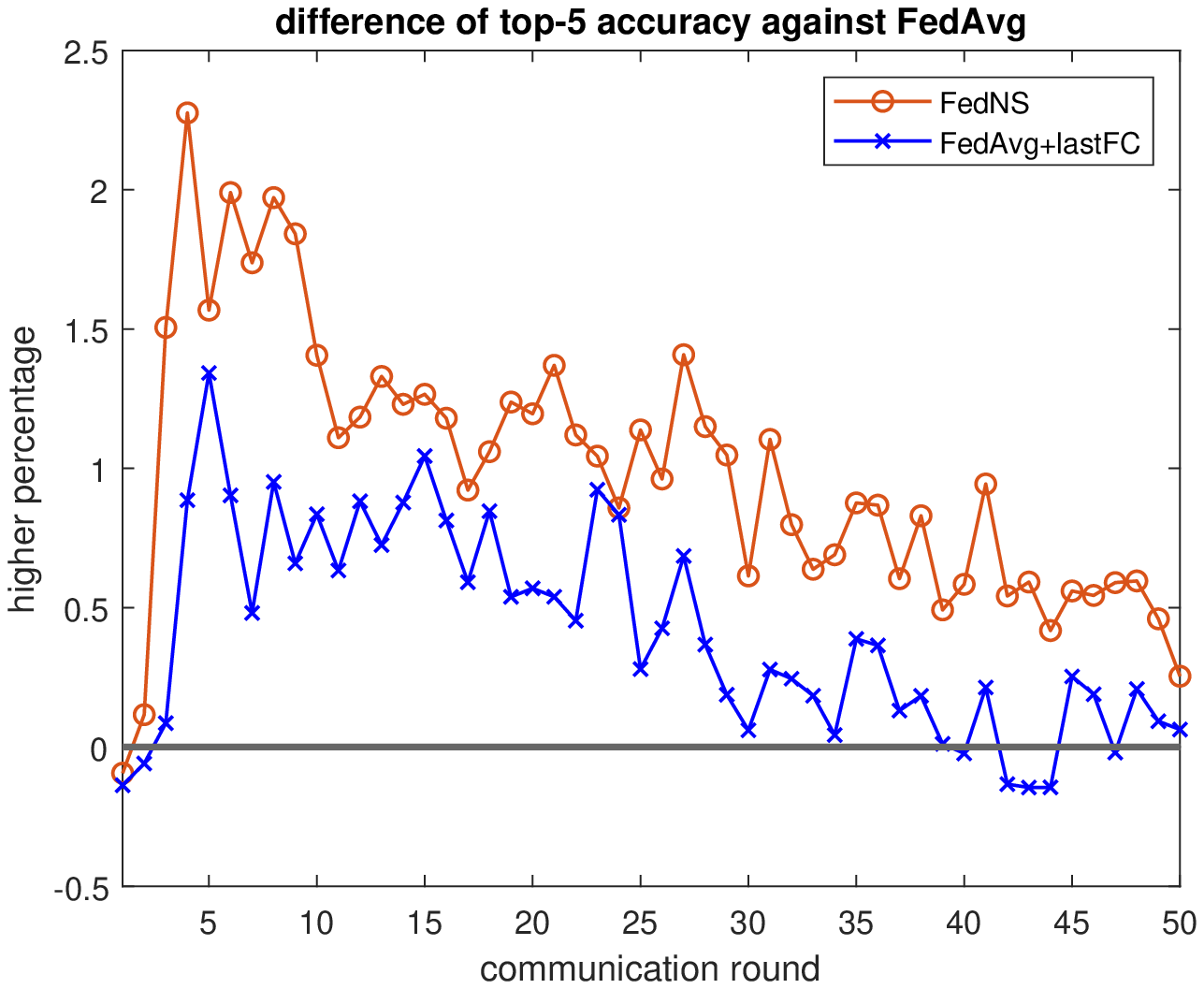,width=4cm}}
      \centerline{non-iid top-5 accuracy}\medskip
    \end{minipage}
    
    \caption{tinyImageNet iid and non-iid daily comparison}
    \label{fig:tinyImageNet daily}
\end{figure}

For each dataset and approach, we repeat the experiment ten times and report the averaged results. It is worth noting that FedAvg+lastFC and FedAvg are the same in the iid condition, and thus we only compare FedNS with FedAvg in this case. Figure \ref{fig:FashionMNIST daily} and Table \ref{tab:final-FashionMNIST} show the results of FashionMNIST with 5-layer CNN. Figure \ref{fig:CIFAR10 daily} and Table \ref{tab:final-CIFAR10} show the results of CIFAR10 with AlexNet. Figure \ref{fig:tinyImageNet daily} and Table \ref{tab:final-tinyImagenet} show the results of tinyImageNet with ResNet18. The tables show the final global model's performance and figures show the progressive performance differences for the methods in each communication round. From the figures and tables, we observe that overall FedNS outperforms the other two methods, especially in the non-iid conditions. 
Please refer to the supplemental materials for higher-resolution versions of the plots.

\begin{table}[t]
\begin{center}
\caption{The final performance comparison in FashionMNIST dataset with 5-layer CNN in iid and no-idd conditions} \label{tab:final-FashionMNIST}

\resizebox{\columnwidth}{!}{%
\begin{tabular}{c|l|c|c|c|c|c|}
\hline
\multicolumn{2}{|c|}{\multirow{2}{*}{\begin{tabular}[c]{@{}c@{}}Fashion\\ MNIST\end{tabular}}} & \multicolumn{2}{c|}{iid} & \multicolumn{3}{c|}{non-iid}                                                        \\ \cline{3-7} 
\multicolumn{2}{|c|}{}  & FedAvg  & FedNS  & FedAvg & \begin{tabular}[c]{@{}c@{}}FedAvg\\ +lastFC\end{tabular} & FedNS          \\ \hline
\multicolumn{2}{|c|}{accuracy}& 83.53   & \textbf{83.79} & 82.96  & 83.50 & \textbf{83.56} \\ \hline
\multicolumn{2}{|c|}{\begin{tabular}[c]{@{}c@{}}macro\\ precision\end{tabular}}  & 0.834   & \textbf{0.838} & 0.830  & 0.837 & \textbf{0.838} \\ \hline
\multicolumn{2}{|c|}{\begin{tabular}[c]{@{}c@{}}macro \\ F-Score\end{tabular}}  & 0.931   & \textbf{0.936} & 0.926  & \textbf{0.933} & 0.929 \\ \hline
\end{tabular}%
}

\end{center}
\end{table}

\begin{table}[t]
\begin{center}
\caption{The final performance comparison in CIFAR10 dataset with AlexNet in iid and non-iid condition} \label{tab:final-CIFAR10}

\resizebox{\columnwidth}{!}{%
\begin{tabular}{|c|l|c|c|c|c|c|}
\hline
\multicolumn{2}{|c|}{\multirow{2}{*}{CIFAR10}}   & \multicolumn{2}{c|}{iid} & \multicolumn{3}{c|}{non-iid}                                                        \\ \cline{3-7} 
\multicolumn{2}{|c|}{}  & FedAvg  & FedNS  & FedAvg & \begin{tabular}[c]{@{}c@{}}FedAvg\\ +lastFC\end{tabular} & FedNS          \\ \hline
\multicolumn{2}{|c|}{accuracy}& 66.09   & \textbf{67.07} & 65.86  & 65.57 & \textbf{67.55} \\ \hline
\multicolumn{2}{|c|}{\begin{tabular}[c]{@{}c@{}}macro\\ precision\end{tabular}}  & 0.658   & \textbf{0.668} & 0.660  & 0.658 & \textbf{0.671} \\ \hline
\multicolumn{2}{|c|}{\begin{tabular}[c]{@{}c@{}}macro \\ F-Score\end{tabular}}  & 0.740   & \textbf{0.745} & 0.737  & 0.737 & \textbf{0.745} \\ \hline
\end{tabular}%
}

\end{center}
\end{table}

\begin{table}[t]
\begin{center}
\caption{The final performance comparison in tinyImageNet dataset with ResNet18 in iid and no-idd conditions} \label{tab:final-tinyImagenet}

\resizebox{\columnwidth}{!}{%
\begin{tabular}{|c|l|c|c|c|c|c|}
\hline
\multicolumn{2}{|c|}{\multirow{2}{*}{\begin{tabular}[c]{@{}c@{}}tiny\\ ImageNet\end{tabular}}} & \multicolumn{2}{c|}{iid} & \multicolumn{3}{c|}{non-iid}                                                        \\ \cline{3-7} 
\multicolumn{2}{|c|}{}  & FedAvg  & FedNS  & FedAvg & \begin{tabular}[c]{@{}c@{}}FedAvg\\ +lastFC\end{tabular} & FedNS          \\ \hline
\multicolumn{2}{|c|}{\begin{tabular}[c]{@{}c@{}}top-1\\ accuracy\end{tabular}}  & 18.99  & \textbf{19.11} & 18.64  & 18.89 & \textbf{18.91} \\ \hline
\multicolumn{2}{|c|}{\begin{tabular}[c]{@{}c@{}}top-5 \\ accuracy\end{tabular}}  & 41.14   & \textbf{41.39} & 40.80  & 40.86 & \textbf{41.05} \\ \hline
\end{tabular}%
}

\end{center}
\end{table}

\section{Conclusion and Future Work}
\label{section conclusion}
 We proposed FedNS as a new model aggregation method for FL. Without much modification to the FL framework, our method can be easily applied to applications like mobile image classification. Different from the traditional FL setting where each client has a fixed fraction of the data, our approach allows flexible and varying data availability, making it closer to the real-world scenario. Even under such challenging settings, experiments showed that FedNS improves upon the baselines, both during the process of learning and at the end of learning. 
One future direction may be to explore how to extend FedNS to 
handle extreme data distributions, such as when each client has distinct classes. In such cases, a more subtle question of comparing models learned from distinct classes needs to be addressed. 


\bibliographystyle{IEEEbib}
\bibliography{icme2021template.bbl}

\end{document}